\documentclass[11pt]{article}
\usepackage{eacl2017}
\usepackage{times}
\usepackage{url}
\usepackage{latexsym}
\usepackage{times}
\usepackage{url}
\usepackage{latexsym}
\usepackage{makecell}
\usepackage{enumerate}
\usepackage{algorithm}
\usepackage{algorithmic}
\usepackage{graphics}
\usepackage{subfigure}
\usepackage{url}
\usepackage{color}
\usepackage{epsfig}
\usepackage{amssymb}
\usepackage{amsmath}
\usepackage{amsfonts}
\usepackage{verbatim}
\usepackage{multirow}
\usepackage{color}
\usepackage{pgfplots}

\eaclfinalcopy 

\newcommand{\red}[1]{{\color{black} #1}}

\newcommand{\eg}{e.g.,~}

\newcommand{\Ni}{({\em i})~}
\newcommand{\Nii}{({\em ii})~}
\newcommand{\Niii}{({\em iii})~}

\newcommand{\good}{\texttt{good}\,}
\newcommand{\bad}{\texttt{bad}\,}
\newcommand{\pot}{\texttt{potentially useful}\,}

\newcommand{\perf}{\texttt{perfect match}\,}
\newcommand{\rel}{\texttt{relevant}\,}
\newcommand{\irel}{\texttt{irrelevant}\,}

\title{Multitask Learning with Deep Neural Networks \\ for Community Question Answering	}

\author{Daniele Bonadiman$^{\dagger}$ \and Antonio Uva$^{\dagger}$ \and Alessandro Moschitti  \\
 $^{\dagger}$DISI, University of Trento, 38123 Povo (TN), Italy \\
 Qatar Computing Research Institute, HBKU, 34110,  Doha, Qatar \\ 
\url{{d.bonadiman, antonio.uva}@unitn.it}\\ \url{amoschitti@gmail.com}
}

\date{}

\begin{document}
\maketitle





\begin{abstract}


In this paper, we developed a deep neural network (DNN) that learns to solve simultaneously the three tasks of the cQA challenge proposed by the SemEval-2016 Task 3, i.e., question-comment similarity, question-question similarity and new question-comment similarity. The latter is the main 
task, which can exploit the previous two for achieving better results. Our DNN is trained jointly on all the three cQA tasks and learns to encode questions and comments into a single vector representation shared across the multiple tasks. The results on the official challenge test set show that our approach produces higher accuracy and faster convergence rates than the individual neural networks. Additionally, our method, which does not use any manual feature engineering, approaches the state of the art established with methods that make heavy use of it.


\end{abstract}



\section{Introduction}

Community Question Answering (cQA) websites enable users to freely ask questions in web forums and expect some good answers in the form of comments from the other users. Given the large number of question/answer pairs available on cQA sites, researchers started to investigate the possibility to exploit user-generated content for training automatic QA systems. Unfortunately, the text involved in the cQA scenario is rather noisy, therefore, providing models that outperform the simple bag-of-words representation can result rather difficult. The challenge, SemEval-2016 Task 3 ``Community Question Answering", has been designed to study the above problems: the participants were supposed to build a fully automatic system for cQA. In particular, given a fresh user question, $q_{new}$, and a set of forum questions, $Q$, answered by a comment set, $C$, the main task consists of determining whether a comment $c \in C$ is a pertinent answer of $q_{new}$ or not. This task can be divided into three sub-tasks: \vspace{-.5em}
\begin{itemize}
  \setlength\itemsep{-.3em}
\item 
[(A)] predict if a comment produced in response to a question contains a valid answer; 
\item 
[(B)] re-rank a set of questions according to their relevancy with respect to the original question; and 
\item 
[(C)] predict if a comment produced in response to a previous question posed on the cQA forum represents a valid answer to a fresh question.
 \vspace{-1.2em}
 \end{itemize}


Traditionally, these tasks have been tackled by designing systems/classifiers that target each task separately. Each classifier accepts in input a vector encoding  a text pair (e.g., a question/question or a question/answer pair)  by using many complex lexical syntactic or semantic features and, then, computing similarity between these representations. However, this approach suffers from the drawbacks of requiring a ``customized'' set of features for each task being solved. 

Recent work on deep neural networks (DNNs) for Multitask Learning (MTL)  \cite{collobert2008unified,liu2015representation} showed that is possible to {\em jointly train} a general system that solves different tasks simultaneously. 
%
%
Inspired by the success of MTL, in this paper, we propose a DNN model that leverages the data from the three cQA tasks of SemEval. Indeed, as the three tasks  are highly related, we claim  that cQA  can highly benefit from this approach. We show that our DNN, despite the fact that does not require any feature engineering, approaches the performance of the best systems, which use heavy feature engineering. Additionally, we are going to make the corpora for studying MTL on this interesting challenge available to the research community.

\section{ cQA Tasks at SemEval}

\label{tasks}
\begin{figure}[t]
    \begin{center}
    
    \includegraphics[scale=0.52]{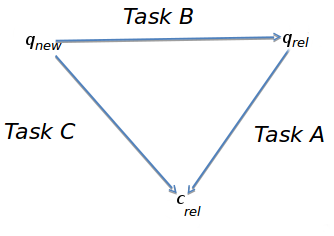}
    \end{center}
    \label{fig:triangolo-cqa}
    \vspace{-1em}
\caption{The 3 tasks of cQA at SemEval: the arrows show the relations between the original and the related questions and the related comments.
}
    
\end{figure}

The research problem issued by SemEval-2016 Task 3 is exemplified by Fig.~\ref{fig:triangolo-cqa}: given a new question $q_{new}$, Task C is about directly retrieving a relevant comment from the entire community. This can also be achieved by solving Task B, which finds a similar question, $q_{rel}$, and then executing Task A, which selects good comments, $c_{rel}$, for $q_{rel}$. It should be noted that Task A classifies comments, specifically written by the users for $q_{rel}$, whereas Task C classifies comments written by the users for other, sometimes, similar questions. This means, it needs to filter out comments that can be partially related to $q_{new}$ (because they correctly answer the related question, $q_{rel}$) but still not correctly answering $q_{new}$. Clearly, Task C classifier needs to tackle a much more semantically challenging task.
Thus, tasks A and C are semantically and computationally rather different and together with Task B, they constitute an interesting MTL problem since differences and correlations are played at a very high semantic level.


\subsection{Task A: Question-Comment Similarity}

Given a question, $q_{rel}$, and its first 10 comments, $c_{rel}$, in the question threads, rerank the comments according to their relevance to $q_{rel}$. Relevancy is defined according to three classes: \Ni \good: the comment is definitively relevant; 
\Nii \pot: the comment is not good, but it still contains related information worth checking; and 
\Niii \bad: the comment is irrelevant (\eg it is part of a dialogue or unrelated 
to the topic). For evaluation purposes, both \pot and \bad \ comments were considered as \bad.

\subsection{Task B: Question-Question Similarity}
Given a new question, $q_{new}$, and its first 10 related questions  (retrieved by a search engine), $q_{rel}$, rerank them according to their similarity with respect to $q_{new}$. Relevancy is expressed by three classes: \Ni \perf: the new and forum questions request roughly the same information,
\Nii \rel: the new and forum questions ask for similar information, or 
\Niii \irel: the new and forum questions are completely unrelated.  
For evaluation purposes, both \perf and \rel forum questions are considered as \rel.

\subsection{Task C: New Question-Comment Similarity}
\label{sect:taskc}

Given a new question, $q_{new}$, and its first 10 related questions  (retrieved by a search engine), $q_{rel}$, each associated with its first 10 comments, $c_{rel}$, appearing in its thread, rerank the 100 comments (10 questions $\times$ 10 comments) according to their relevance with respect to $q_{new}$. Relevancy is defined similarly to task A.

\subsection{Dataset}

The data for the above-mentioned tasks is distributed in three datasets: train, dev and test sets. The distribution of questions and comments in each dataset varies across the different tasks: Task A contains 6,938 related questions and 40,288 comments. Each comment in the dataset was annotated with a label indicating its relevancy  with respect to the related question.  Task B contains 317 original questions. For each original question, 10 related questions were retrieved, summing to 3,169  related questions. Also in this case, the related questions were annotated with a relevancy label, which tells if they are relevant with respect to the user original question. 
Task C contains 317 original question, together with 3,169 related questions (same as in Task B) and 31,690 comments.  Each comment was labelled with its relevancy with respect to the original question. 

\section{A General Deep Architecture for cQA}

\label{sec:gdcqa}

All the previous tasks are about reranking questions or comments with respect to an original question. In the following, we describe a general architecture for solving them.

\subsection{Deep Architecture for relational learning from pairs of text}

\label{sec:pair_arch}
A traditional approach to cQA is to learn a different classifiers for solving each of these three tasks, independently. For example, first a classifier can be trained to rerank a set of related questions retrieved by a search engine, using their similarity with respect to the user question (Task B). Then, another classifier can be trained to rerank the list of comments appearing in the threads of similar questions (Task A). Each of these classifiers uses a different set of task-dependent features. 
In this work, we use a neural network architecture for classifying text pairs.  The network is fed using the different pairs, ($q_{rel}$, $c_{rel}$), ($q_{new}$, $q_{rel}$) and ($q_{new}$, $c_{rel}$), for learning the tasks A, B and C, respectively, and produces a similarity score that can be used for reranking questions or comments.

 %
%
It is composed of two main components: (i) two sentence encoders that map input sentences $i$ into fixed size vectors $x_{s_i} \in \mathbb{R}^{m}$, and (ii) a feed forward neural network that computes the similarity between these two sentence vectors.

The sentence encoders are composed of (i) a sentence matrix $\mathbf{s}_{i} \in \mathbb{R}^{d \times |\mathbf{i}|}$, where $d$ is the size of the word embeddings, obtained by concatenating the word vector of the corresponding word in the input sentence $\mathbf{w}_{j} \in \mathbf{s}_{i}$, and (ii) a sentence model $f: \mathbb{R}^{d \times |\mathbf{i}|} \rightarrow \mathbb{R}^m$, which maps the sentence matrix to a fixed size sentence embedding  $x_{s_i} \in \mathbb{R}^m$.

The choice of the sentence model plays a crucial role as the resulting intermediate representation of the input sentences affects the successive steps of computing their similarity. Previous work in this direction uses different types of sentence models such as LSTM, distributional sentence model (average of word vectors), and convolutional sentence model. In particular, the latter is composed of a sequence of convolution and pooling feature maps have achieved the state of the art in various NLP tasks~\cite{nal:2014,kim:2014}.
	
In this paper, we used a CNN sentence model that is a convolutional operation followed by a $k$-max pooling layer with $k=1$, since it provides comparable performance to the LSTM on the task of new question-comment similarity, as shown in Table~\ref{table:cqa-exp-nn}.
The sentence encoder, $x_{s_{i}} = f(s_{i})$, outputs a fixed-size embedding of the input sentence $s_{i}$. 
The sentence vectors, $x_{s_i}$, are concatenated together and given in input to a Multi-Layer Perceptron, which is constituted by a non-linear hidden layer and an sigmoid output layer. 


\subsection{Injecting Relational Information}
\label{INJ}

All the tasks we consider require to model 
relations between words present in the two pieces of text. 
For this purpose, we encode the relation in forms of discrete features, as described in~\cite{collobert2011natural}, i.e., using an additional embedding layer. They augmented the word embedding with the corresponding feature embedding. Thus, given a word, $w_j$, its final word embedding is defined as $\mathbf{w}_j \in \mathbb{R}^d$, where $d = d_w + d_{feat}$, where $d_w$ is the size of the word embedding and $d_{feat}$ the size of the feature embedding. 
%


We use a discrete feature, represented with an embedding of 5 dimensions,  to encode matches between two words in the two input piece of text. 
In particular, we associate each word $w$ in the input sentences with a \textit{word overlap} index $o \in \{0, 1\}$, where $o=1$ means that $w$ is shared by both Q and C (or by the two questions for task B), i.e., overlaps, $o=0$ otherwise. 
It should be noted that the embeddings described here cannot be considered as task specific features, manually handcrafted. They are part of the network, serve the purpose of injecting relational information between the representations of the two input texts and can be generally applied to different domains, data and tasks.

\subsection{Adding the rank features}

\label{adapt-taskc}
The SemEval problems concern reranking text initially ranked by Google and made available to the participants for tasks B and C. Considering that the Google rank is computed using powerful algorithms and a lot of resources, it is essential to encode it in our networks.  There are several methods for achieving this. After some experiments, we opted for discretizing the rank values in 5 different bins of different sizes, i.e. $[1-2], [2-5], [5-10], [10-25], [25-\infty]$. The rank feature is added to the joint layer, where the output of the sentence model is concatenated, using a table  lookup operation.
It should be noted that for each task, we use a different relation feature (overlap embeddings) between each pair of text.

\section{MTL for cQA}

\label{sec:multitask}
\begin{figure*}[t]
\center
\includegraphics[scale=0.60]{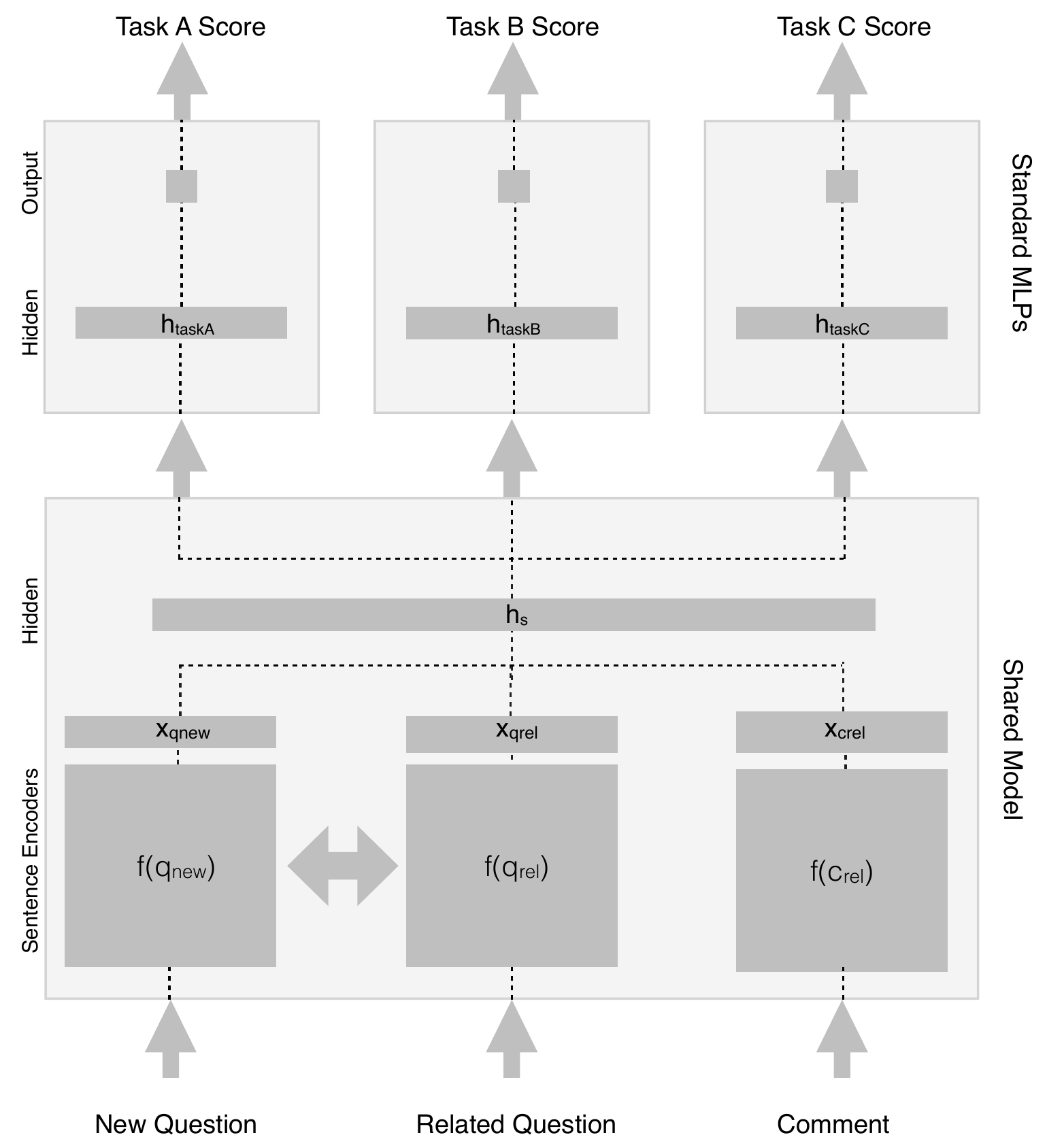}

\caption{\small Our MTL architecture, where the three sentences are at the bottom. They pass through the sentence encoders. The output is concatenated and fed to a hidden layer whose output is passed to three independent multi-layer perceptrons, which produce the scores for the individual tasks.  The $\leftrightarrow$ symbol indicates a shared sentence model between $q_{new}$ and $q_{rel}$.}
\label{fig:mtl}

\end{figure*}

MTL aims at learning several related tasks at the same time for improving  some (or possibly all) tasks using joint information \cite{caruana1997multitask}.  
MTL is particularly well suited for modeling Task C as it is a composition of tasks A and B, thus, it can benefit from having both questions $q_{new}$ and $q_{rel}$ as input to better model the interaction between the new question and the comment. More precisely, it can use the triplet $\langle q_{new}, q_{rel}, c_{rel} \rangle$ in the learning process, where the interaction between the triplet members is exploited during the joint training of the three models for tasks, A, B and C.
%
In fact, an improvement on question-comment similarity or on question-question similarity can lead to an improvement in the task of new question-comment similarity (Task C). 

Additionally, each thread in the in the SemEval dataset is annotated with the labels for all the three tasks and therefore it is possible to apply joint learning directly. 



\subsection{Joint Learning Architecture}

Our Joint learning architecture is depicted in Figure~\ref{fig:mtl}, it is a direct extension of the architecture proposed for Task C (Section~\ref{adapt-taskc}). It takes the three sentences as input, i.e, a new question, $q_{new}$, the related question, $q_{rel}$, and its comment, $c_{rel}$, and produces three fixed size representations, $x_{q_{new}}$, $x_{q_{rel}}$ and $x_{c_{rel}}$, respectively.

These three representations are then concatenated ($h_{j} = [x_{q_{new}},x_{q_{rel}},x_{r_{rel}}]$) and fed to a hidden layer to create a shared representation of the input for the three tasks, $h_{s} = Wh_{j}$.

The output of this layer, $h_{s}$ is then fed to three independent Multilayer Perceptrons (MLP) that produce the scores for the three tasks. 
To directly apply MTL, we use the binary cross-entropy instead of the max margin loss as our objective function. The main motivation is that such function is computed based on pairs of positive-negative examples that cannot be created with multiple labels. At training time, for each example, the loss is calculated on the three outputs of the network. The final loss is then the sum of the individual losses for the three tasks.

\subsection{Shared Sentence Models}
\label{ssec:ssm}
The SemEval dataset contains ten times less new questions than related questions by construction. However, all questions, $q_{new}$ included, are supposed to be of the same nature. Thus we can certainly use a shared text model for modeling better representations for both new and related questions. Formally, let $x_d = f(d, \theta)$ be a sentence model for document $d$ with parameters $\theta$, i.e., the embedding weights and the convolutional filters. In our original formulation, each sentence model uses a different set of parameters $\theta_{q_{new}}$, $\theta_{q_{rel}}$ and $\theta_{c_{rel}}$. We used the same set of parameters $\theta_{q}$. The shared sentence model is depicted in Figure~\ref{fig:mtl} as $\leftrightarrow$.

\begin{table}[]
\vspace{.5em}
\centering
\small
\begin{tabular}{|c|c|c|c|}
\hline
                        & Task A  & Task B  & Task C   \\ \hline
Train                   & 37.51\% & 39.41\% & 9.9\%          \\
Train + ED & 37.47\% & 64.38\% & 21.25\%      \\ \hline
\end{tabular}
\caption{Percentage of positive examples in the training datasets for each task.}
\vspace{-1em}
\label{data-distribution}
\end{table}

\section{Experiments}
\label{EXP}
 
\subsection{Setup} 

We encode input sentences with fixed-sized vectors using a convolutional operation of size 5 and a $k$-max pooling operation with $k = 1$, i.e., similarly to \cite{severyn2015learning}. We use two non-linear hidden layers (with hyperbolic tangent activation, Tanh), whose size is equal to the size of the previous layer, i.e., the join layer. We include information such as word overlaps and rank position as embedding with an additional lookup table with vectors of size $d_{feat} = 5$.

{\bf Pre-processing}: both questions and comments are tokenized and lowercased (to reduce the dimensionality of the dictionary and therefore of the embedding matrix).
Moreover, question subject and body are concatenated to create a unique question. 
For computational reasons, we opted to limit the size of the input documents at 100 words: we did not observe any degradation in performance.

{\bf Word Embeddings}:  for all the proposed models, we pre-initialize the word embedding matrices with standard skipgram embedding of dimensionality 50 trained on the English Wikipedia dump using word2vec toolkit~\cite{word2vec}.

{\bf Training}:  The network is trained using SGD with shuffled mini-batches using the rmsprop update rule~\cite{tieleman2012lecture}. The model learns until the validation loss stops improving, with patience $p=10$, i.e., the number of epochs to wait before early stopping, if no progress on the validation set is obtained. In fact, early stopping~\cite{prechelt1998early} allows us to avoid overfitting and improving the generalization capabilities of the network. For the MTL architecture, we employed two different stopping criteria. The first is to stop training when the global validation loss does not improve anymore (the sum of the individual losses of the three tasks). The second, instead, saves three different models and evaluates them when the individual losses stop improving. Since the three tasks converge at different epochs, the first method may lead to sub-optimal results for the individual tasks, but only one model is needed at test time.

To improve generalization and avoid co-adaptation of features, we opted for adding dropout~\cite{srivastava2014dropout} between all the layers of the network. We experimented with different dropout rates (0.2, 0.4) for the inputs and (0.3, 0.5, 0.7) for the hidden layers obtaining better results with the highest values, i.e., 0.4 and 0.7.

{\bf Dataset}: Table~\ref{data-distribution} reports the labels distributions on the train dataset. It is important to note that the dataset for Task C presents a higher number of negative than positive examples.
For this reason, we automatically extended the training dataset (ED) with new positive matches for Task B and C, respectively. This process is done by creating the $(q_{rel},c_{rel})$ pairs for each $q_{rel}$ from the training set for Task A and creating triples of the form $(q_{rel}, q_{rel}, c_{rel})$, where the label for question-question similarity is obviously positive and the labels for Task C are inherited  from those of Task A. The resulting dataset contains $34,100$ triples and its relevance label distribution is presented in the last row of Table~\ref{data-distribution}. The extended version of the dataset with the annotation for MTL is made available for download for comparison purposes~\footnote{Download link to be defined.}.

\begin{table}[t]
\small

\hspace{3em}
\center
\begin{tabular}{|l|l|l|}
\hline
Model & MAP & MRR\\

\hline
\hline

LSTM&43.91&49.28\\
CNN&44.43&49.01\\
\hline
\hline
CNN Train&44.43&49.01\\
CNN Train + ED\footnotemark&\textbf{44.77}&\textbf{52.07}\\
\hline
\end{tabular}
\caption{Impact of CNN vs.~LSTM sentence models on the baseline network for Task C.}
\label{table:cqa-exp-nn}
\vspace{-1em}
\end{table}
\footnotetext{Extended Dataset for Task C computed using questions from Task A. }



\begin{table*}[t]
\center
\small
\scalebox{0.9}{
\begin{tabular}{|l|r|r|r|r|r|r|r|r|r|r|r|c|}
\hline 
\multirow{3}{*}{Models}&\multicolumn{4}{|c|}{\makecell{Task A: \\ question-comment similarity}}&\multicolumn{4}{|c|}{\makecell{Task B: \\ question-question similarity}}&\multicolumn{4}{|c|}{\makecell{Task C: \\ new question-comment similarity}}\\ \cline{2-13} 
&\multicolumn{2}{|c|}{DEV}&\multicolumn{2}{|c|}{TEST}&\multicolumn{2}{|c|}{DEV}&\multicolumn{2}{|c|}{TEST}&\multicolumn{2}{|c|}{DEV}&\multicolumn{2}{|c|}{TEST}\\  \cline{2-13} 
&MAP&MRR&MAP&MRR&MAP&MRR&MAP&MRR&MAP&MRR&MAP&MRR\\
\hline
\hline
Random&-&-&59.53&67.83&-&-&46.98&50.96&-&-&15.01&15.19\\
IR Baseline&-&-&52.80&58.71&-&-&74.75&83.79&-&-&40.36&45.83\\
\hline
\hline
Kelp&-&-&79.19&86.42&-&-&-&-&-&-&-&-\\								
UH-PRHLT&-&-&-&-&-&-&76.70&83.02&-&-&-&-\\				
SUper-team&-&-&-&-&-&-&-&-&-&-&55.41&61.48\\
\hline
\hline
$\langle q_{rel}, c_{rel} \rangle$ &68.93&76.46&74.73&81.18&-&-&-&-&-&-&-&-\\
$\langle q_{new}, q_{rel} \rangle$ &-&-&-&-&74.19&\textbf{83.26}&\textbf{73.70}&\textbf{82.13}&-&-&-&-\\
$\langle q_{new}, c_{rel} \rangle$ &-&-&-&-&-&-&-&-&44.77&52.07&41.95&47.21\\
\hline
\hline
$\langle q_{new}, q_{rel}, c_{rel} \rangle$ &-&-&-&-&-&-&-&-&45.59&51.04&46.99&55.64\\
$\langle q_{new}, q_{rel}, c_{rel} \rangle$ + $\leftrightarrow$ &70.69&77.19&\textbf{75.52}&82.11&72.92&80.20&72.88&80.58&47.82&53.03&46.45&51.72\\
\hline
\hline
MTL (BC)&-&-&-&-&\textbf{74.22}&80.40&73.68&81.59&47.80&52.31&48.58&\textbf{55.77}\\
MTL (AC)&70.11&76.50&75.43&\textbf{82.46}&-&-&-&-&46.34&51.54&48.49&54.01\\
MTL (ABC)&69.93&76.27&74.42&81.68&70.68&75.85&71.07&80.11&\textbf{49.63}&\textbf{55.47}&49.87&55.73\\
MTL (ABC)*&\textbf{70.70}&\textbf{77.48}&74.89&81.80&74.21&81.93&72.23&80.33&\textbf{49.63}&\textbf{55.47}&49.87&55.73\\
\hline
\hline
\red{MTL (weighted score)} &-&-&-&-&-&-&-&-&-&-&\textbf{52.67}&55.68\\
\hline
\end{tabular}}
\caption{Results on the validation and test set for the proposed models}
\label{tab:semevalresult}

\end{table*}

\subsection{Impact of the sentence models}
\label{sec:results}

Table~\ref{table:cqa-exp-nn} shows a comparison between CNN and LSTM sentence models when used in our general architecture (see Sec.~\ref{sec:gdcqa}) for solving Task C.
We derived the results from the development set~\footnote{In this work, the dataset Train-part2 were used as development set.}. 
We observe that the two sentence models show comparable results. 
For the rest of the experiments, we used the CNN sentence model, since it shows faster convergence rate and more stable results with respect to the LSTM sentence model. In the second part of Table~\ref{table:cqa-exp-nn}, we demonstrate that using the extended dataset for solving Task C leads to higher results than the original one. In particular, we noted that there is an improvement of 3 points in MRR.

\subsection{Results of individual models}
Table~\ref{tab:semevalresult} shows the results of our individual and MTL models, in comparison with the Random and Information Retrieval baselines of the challenge (first grouped row), and the three-top systems of SemEval 2016, Kelp, UH-PRHLT, SUper-team (second grouped row).

The third grouped row shows the performance of the individual models when trained on input pairs, $\langle q_{rel}, c_{rel} \rangle$, $\langle q_{new}, q_{rel} \rangle$ and $\langle q_{new}, c_{rel} \rangle$  for task A, B and C, respectively. The model for the three tasks is the same (described in Sec.~\ref{sec:gdcqa}). These results show that the individual models can generalize well enough on all tasks.	 In particular, on Task B, they achieve the best results of our proposed model (the numbers in bold indicate the best results among the proposed models).

The fourth grouped row illustrates the models exploiting the joint input, $\langle q_{new}, q_{rel}, c_{rel} \rangle$, but no joint learning is carried out, i.e., the networks for the different tasks are trained individually.
The results show that a small degradation of performance happens in Task B, while Task A slightly improves. These variations may be due to the fact that tasks A and B can be efficiently solved using the standard pairwise approach, thus the extra text introduced in the model may just add some noise.
However, using the shared sentence model for $q_{new}$ and  $q_{rel}$ of the tasks B and C (indicated with $\leftrightarrow$)  improves the overall performance.



\vspace{-.5em}
\subsection{Results of MTL models}
The shared input representation shows good results on all tasks, thus, in the last set of experiments, we jointly trained (i) tasks B and C, (ii) tasks A and C and finally (iii) the three tasks together. 

The results are reported in the fifth grouped row. It is interesting to note that the major boost in terms of performance is obtained when we jointly train all the three tasks. In fact, the MTL architecture improves the individual model in terms of MAP by about $2$ absolute points on the DEV set and by $3$ absolute points on the TEST set for Task C, while the performance on the other tasks tends to degrade. However, if the three different models are evaluated at different epochs of training, e.g., see MTL(ABC)*, it is possible to obtain accuracy comparable to the individual models for all the three tasks. As previously explained, when applying MTL, the individual objective functions converge at different epochs. 
Therefore, when the global loss reaches the minimum, it is possible that individual models are sub-optimal.

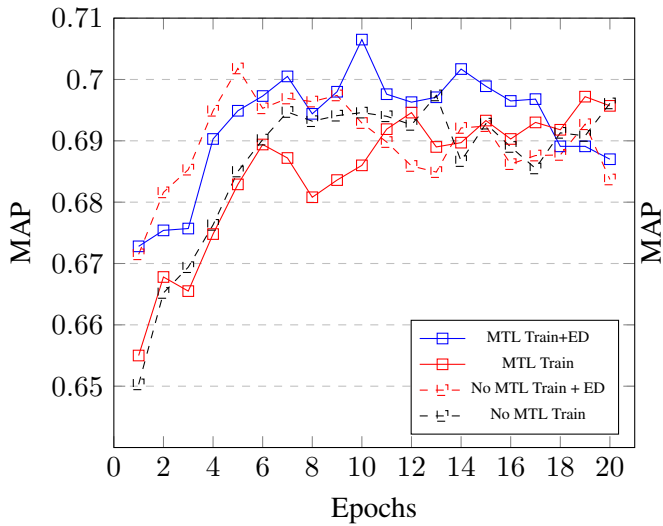
\begin{figure}[t]
\hspace{-1.9em}
\begin{tikzpicture}

\begin{axis}[
    xlabel={ Epochs },
    ylabel={ MAP },
    xmin=0, xmax=21,
    ymin=0.64, ymax=0.71,
    xtick={0,2,4,6,8,10,12,14,16,18,20},
    ytick={0.60, 0.65, 0.66, 0.67, 0.68, 0.69, 0.70, 0.71},
      legend style={legend pos=south east,font=\tiny},ymajorgrids=true, grid style=dashed],
]

\addplot[
    color=blue,
    mark=square,
    ]
    coordinates {
(1, 0.6728)(2, 0.6754)(3, 0.6757)(4, 0.6903)(5, 0.6949)(6, 0.6973)(7, 0.7005)(8, 0.6944)(9, 0.698)(10, 0.7065)(11, 0.6976)(12, 0.6963)(13, 0.6971)(14, 0.7017)(15, 0.6989)(16, 0.6965)(17, 0.6968)(18, 0.6891)(19, 0.6891)(20, 0.687)

    };
    \addlegendentry{MTL Train+ED}
 
\addplot[
    color=red,
    mark=square,
    ]
    coordinates {
	(1, 0.655)(2, 0.6678)(3, 0.6655)(4, 0.6748)(5, 0.6829)(6, 0.6894)(7, 0.6872)(8, 0.6808)(9, 0.6836)(10, 0.686)(11, 0.6919)(12, 0.6946)(13, 0.689)(14, 0.6897)(15, 0.6933)(16, 0.6903)(17, 0.693)(18, 0.6918)(19, 0.6972)(20, 0.6957)

    };
     \addlegendentry{MTL Train}
     
     \addplot[
    color=red,
    mark=square,
    style=dashed,
    ]
    coordinates { (1, 0.6715)(2, 0.6816)(3, 0.6853)(4, 0.6949)(5, 0.7018)(6, 0.6953)(7, 0.6969)(8, 0.6964)(9, 0.6974)(10, 0.6929)(11, 0.6898)(12, 0.6859)(13, 0.6849)(14, 0.6921)(15, 0.6924)(16, 0.6862)(17, 0.6876)(18, 0.6877)(19, 0.6927)(20, 0.6837)

    };
     \addlegendentry{No MTL Train + ED}

     \addplot[
    color=black,
    mark=square,
    style=dashed,
    ]
    coordinates { (1, 0.6503)(2, 0.6652)(3, 0.6694)(4, 0.6763)(5, 0.685)(6, 0.6902)(7, 0.6947)(8, 0.6932)(9, 0.6941)(10, 0.6946)(11, 0.694)(12, 0.6926)(13, 0.6974)(14, 0.6867)(15, 0.6928)(16, 0.689)(17, 0.6855)(18, 0.6913)(19, 0.6908)(20, 0.6961)

    };
     \addlegendentry{No MTL Train} 
\end{axis}
\end{tikzpicture}

\vspace{-1em}
\caption{\small Learning curves for Task A on the dev. set; dotted and solid lines represent the individual and multi-task models, respectively.} \label{fig:taska}
\vspace{-.5em}
\end{figure}

\begin{figure}[t]
\hspace{-1.9em}
\begin{tikzpicture}

\begin{axis}[
    xlabel={ Epochs },
    ylabel near ticks,
    ylabel={ MAP },
    xmin=0, xmax=21,
    ymin=0.45, ymax=0.75,
    xtick={0, 2, 4, 6, 8, 10,12,14,16,18,20},
    ytick={0.50,0.55,0.60,0.65,0.70,0.75},
    legend style={legend pos=south east,font=\tiny},ymajorgrids=true, grid style=dashed],
]

\addplot[
    color=blue,
    mark=square,
    ]
    coordinates {
 (1, 0.7202)(2, 0.7392)(3, 0.7146)(4, 0.6834)(5, 0.6897)(6, 0.687)(7, 0.6955)(8, 0.6887)(9, 0.6878)(10, 0.6891)(11, 0.6799)(12, 0.6843)(13, 0.687)(14, 0.674)(15, 0.6844)(16, 0.6857)(17, 0.6785)(18, 0.6949)(19, 0.6932)(20, 0.6936)

    };
    \addlegendentry{MTL Train+ED}
 
\addplot[
    color=red,
    mark=square,
    ]
    coordinates {
	(1, 0.6041)(2, 0.7246)(3, 0.7187)(4, 0.6941)(5, 0.7038)(6, 0.6819)(7, 0.6846)(8, 0.6789)(9, 0.6577)(10, 0.6776)(11, 0.6958)(12, 0.662)(13, 0.6601)(14, 0.6672)(15, 0.6811)(16, 0.6487)(17, 0.6562)(18, 0.6683)(19, 0.6646)(20, 0.6727)

    };
     \addlegendentry{MTL Train}
     
     \addplot[
    color=red,
    mark=square,
    style=dashed,
    ]
    coordinates { (1, 0.6774) (2, 0.6989) (3, 0.6976)(4, 0.7017)(5, 0.702)(6, 0.6986)(7, 0.6978)(8, 0.6922)(9, 0.6929)(10, 0.6962)(11, 0.695)(12, 0.6927)(13, 0.6904)

    };
     \addlegendentry{No MTL Train + ED}

     \addplot[
    color=black,
    mark=square,
    style=dashed,
    ]
    coordinates { (1, 0.526)(2, 0.6367)(3, 0.6556)(4, 0.6787)(5, 0.6901)(6, 0.6482)(7, 
0.6719) (8, 0.6595)(9, 0.677)(10, 0.6735)(11, 0.6706)(12, 0.6769)(13, 0.6704)(14, 0.6727)(15, 0.6693)(16, 0.6656)

    };
     \addlegendentry{No MTL Train} 
\end{axis}
\end{tikzpicture}

\vspace{-1em}
\caption{\small Learning curves for Task B on the development set; dotted lines represent the individual models, while the solid lines the multi-task ones.} \label{fig:taskb}
\end{figure}
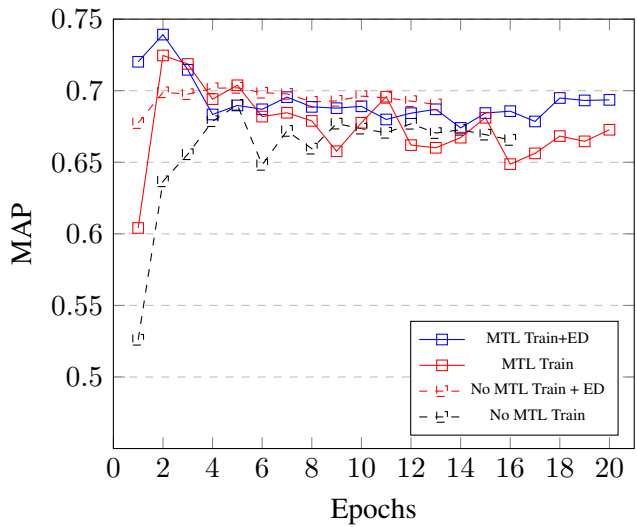

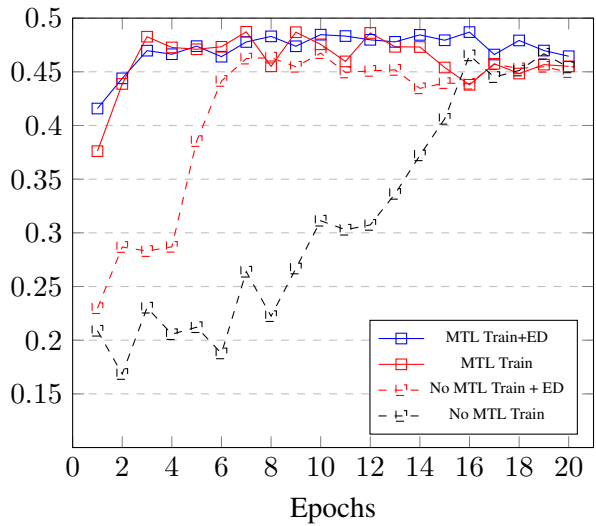
\begin{figure}
\hspace{-2em}
\begin{tikzpicture}

\begin{axis}[
    xlabel={ Epochs },
    ylabel={ MAP },
    xmin=0, xmax=21,
    ymin=0.10, ymax=0.5,
    xtick={0, 2, 4, 6, 8, 10,12,14,16,18,20},
    ytick={0.15,0.20,0.25,0.30,0.35,0.40,0.45,0.50},
    legend style={legend pos=south east,font=\tiny},ymajorgrids=true, grid style=dashed],
]

\addplot[
    color=blue,
    mark=square,
    ]
    coordinates {
 (1, 0.4158)(2, 0.4439)(3, 0.4697)(4, 0.4664)(5, 0.4739)(6, 0.4641)(7, 0.4778)(8, 0.483)(9, 0.4738)(10, 0.4846)(11, 0.4833)(12, 0.4798)(13, 0.4779)(14, 0.4844)(15, 0.4794)(16, 0.487)(17, 0.4661)(18, 0.4792)(19, 0.4698)(20, 0.4645)
    };
    \addlegendentry{MTL Train+ED}
 
\addplot[
    color=red,
    mark=square,
    ]
    coordinates {
(1, 0.3761)(2, 0.4387)(3, 0.4827)(4, 0.4727)(5, 
0.471)(6, 0.4733)(7, 0.4873)(8, 0.4551)(9, 0.4869)(10, 0.4756)(11, 0.4597)(12, 0.4862)(13, 0.4733)(14, 0.4731)(15, 0.454)(16, 0.4379)(17, 0.4575)(18, 0.4484)(19, 0.4567)(20, 0.4551)
    };
     \addlegendentry{MTL Train}
     
     \addplot[
    color=red,
    mark=square,
    style=dashed,
    ]
    coordinates { (1, 0.2298) (2, 0.2868) (3, 0.283) (4, 0.287) (5, 0.3855) (6, 0.4415) (7, 0.4623) (8, 0.4632) (9, 0.4547) (10, 0.4674) (11, 0.4494) (12, 0.4509) (13, 0.4518) (14, 0.4346) (15, 0.4395) (16, 0.4389) (17, 0.457) (18, 0.4532) (19, 0.4547) (20, 0.4497)

    };
     \addlegendentry{No MTL Train + ED}

     \addplot[
    color=black,
    mark=square,
    style=dashed,
    ]
    coordinates { (1, 0.2088)(2, 0.1685) (3, 0.2303)(4, 0.2056)(5, 0.2121)(6, 0.1877)(7, 0.2638)(8, 0.2224)(9, 0.2668)(10, 0.3113)(11, 0.3029)(12, 0.3074)(13, 0.3363)(14, 0.3722)(15, 0.4062)(16, 0.4653)(17, 0.4451)(18, 0.4512)(19, 0.4668)(20, 0.4541)
    };
     \addlegendentry{No MTL Train} 
\end{axis}
\end{tikzpicture}
\vspace{-2em}
\caption{\small Learning curves for Task C on the development set; dotted lines represent the individual models, while the solid lines the multi-task ones.} \label{fig:taskc}
\vspace{-.5em}
\end{figure}

Indeed, the comparison between the learning curves (on the development set) for Task B (Figure~\ref{fig:taskb}) and Task C (Figure~\ref{fig:taskc}) shows that for the former, models achieve earlier convergence rate (epoch 2) while for the latter they converge later (epoch 16).
Moreover, Figure~\ref{fig:taska} shows that the results on Task A are not badly affected by jointly training models with the other two tasks.

Finally, the learning curves show that our networks trained in MTL tend to have faster convergence rate than the individual models: this is a very interesting result.


We also experimented with shallower networks and SVMs using the prediction scores from the different classifiers in a stacking approach, and obtained results far below the baselines\footnote{We did not include these results as they do not provide interesting findings.}.

\paragraph{Comparison with the state of the art}
Our models would have ranked $4^{th}$ on Task C of the Semeval 2016 competition~\footnote{http://alt.qcri.org/semeval2016/task3/index.php?id=results}, i.e., the main task of the challenge. In contrast, our models  for the other two tasks, which do not benefit from the overall MTL architecture would have achieved a middle position ($8^{th}$). These results are important since our proposed MTL architecture obtains a placement very close to the top system, without requiring task-specific features, which in cQA are extremely important, e.g., the thread-level features. 

Finally, one reason of why we do not achieve the state of the art on Task C is due to the difference between training and test data. Several challenge participants solved this problem by using a weighted sum between the score of the Task A classifier and the Google rank as a strong features for modeling Task C. We followed a similar approach estimating the weight MTL on the dev set and using the computed score to rank the comments of the test set. This improved the MAP of our MTL by about 2.8 absolute points on the test set, obtaining a result comparable with the model ranked $2^{nd}$ on Task C at the Semeval 2016 competition.





\section{Related Work}

{\bf Question-Question Similarity. } Determining question similarity remains one of the main tasks need to be solved in  cQA due to difficult problems such as ``lexical gap''. 
Early approaches on question similarity used statistical machine translation techniques to measure  similarity between  questions.
For example,
\newcite{jeon2005finding} and \newcite{zhou2011phrase} used a language models based on word or phrase translation probabilities for estimating similarity between questions. However, effective approaches based on statistical machine translation require lots of data for estimating word probabilities. 
Language models for question-question similarity were also explored by \newcite{cao2009use}. These models exploit information from the category structure of Yahoo!~Answers when computing similarity between two questions.
Instead, \newcite{duan2008searching} propose an approach that identifies the topic and focus in  a text and compute similarity between two input questions by matching the extracted topic and focus information.
 %
A different approach to question-question similarity is provided by \newcite{ji2012question} and \newcite{zhang2014question}. They use LDA  to learn the  probability distribution over the topics that generate the question/answers pairs. Later, this distribution is used to measure similarity between questions.



\noindent {\bf Question-Answer Similarity.} 
In recent years, many models have been proposed for computing similarity of an answer with respect to a question. For example,
\newcite{yao2013answer} trained a conditional random field based on a set of powerful features, such as tree-edit distance between question and answer trees: these also enable the extraction of answers from pre-retrieved sentences. 
\newcite{heilman2010tree} use a linear classifier using syntactic features to solve  different tasks such as recognizing textual entailment, paraphrases and answer selection. 
\newcite{wang2007jeopardy} propose the use of  Quasi-synchronous grammars to select short answers for TREC questions. This is done by learning syntactic and semantic transformation from the question to the answer trees. 
\newcite{wang_manning:acl:2010} propose a probabilistic Tree-Edit model with structured latent variables for solving textual entailment and question answering. 
An advanced model based on structural representations was proposed by \newcite{severyn2012structural}. This model uses SVM with kernels to learn structural patterns between questions and answers encoded in form of shallow syntactic parse trees. 

Finally, \newcite{wang2015long} trained a long short-term memory model for selecting answers to TREC questions. Their model takes words from question and answer sentences as input and returns a score measuring the relevancy of an answer with respect to a given question. A recent work close to ours is \cite{guzman16lluis}, where the authors build a neural network for solving Task A of SemEval. However, this does not approach the problem as MTL.


{\bf Related work on MTL.} 
A good overview on MTL, i.e., learning to solve multiple tasks by using a shared representation with mutual benefit, is given in \cite{caruana1997multitask}. 
\newcite{collobert2008unified} trained a convolutional NN with MTL which, given an input sentence, performs many sequence labeling tasks. 
They showed that jointly training their system on different tasks, such as speech tagging, named entity recognition, etc., significantly improves the performance on the main task, i.e., semantic role labeling, without requiring hand-engineered features.

\newcite{liu2015representation} is the most close work to ours. They used multi-task deep neural networks to map queries and documents into semantic vector representations. This representation is later used into two tasks: query classification and question-answer reranking. Results showed a competitive gain over strong baselines.
In our work, we have presented an architecture that can also exploit joint representation of question and comments, given the strong interdependencies among the different SemEval Tasks.

    

\section{Conclusion}

\label{sec:length}
In this paper we proposed several Deep Neural Networks for the task of automatic cQA. Our main result is a network that can effectively exploit the characteristics of the cQA task for carrying out interesting MTL.
Our network designed and trained in an MTL setting shows better results and a higher convergence rate than individual models that are trained independently. These results are competitive with those of the models participating at the SemEval 2016 competition for cQA.
It should be noted that all the other challenge systems use domain specific features, which are both very important but also rather costly to engineer.

In the future, we would like to use more effective features and combine them with other machine learning methods.

\section*{Acknowledgements}

This work has been partially supported by the EC
project CogNet, 671625 (H2020-ICT-2014-2, Research
and Innovation action) and by the {\sc Iyas} project, Interactive sYstems for Answer Search.

\bibliography{eacl2017}
\bibliographystyle{eacl2017}

\end{document}